# Image retrieval system base on EMD similarity measure and S-tree

Thanh Manh Le[1], Thanh The Van[2]

[1] Hue University, Thua Thien Hue, Vietnam
[2] Center for Information Technology, HoChiMinh city University of Food Industry, HoChiMinh city, Vietnam

**Abstract.** The paper approaches the binary signature for each image based on the percentage of the pixels in each color images, at the same time the paper builds a similar measure between images based on EMD (Earth Mover's Distance). Besides, the paper proceeded to create the S-tree based on the similar measure EMD to store the image's binary signatures to quickly query image signature data. From there, the paper build an image retrieval algorithm and CBIR (Content-Based Image Retrieval) based on a simlilar measure EMD and S-tree. Based on this theory, the paper proceeded to build application and experimental assessment of the process of querying image on the database system which have over 10,000 images.

**Keywords:** CBIR, Image Retrieval, EMD, S-tree, Signature, Signature Tree.

## 1 Introduction

Finding images in a large DATABASE of digital images is a difficult problem, the puzzle is to query the images in a large image DATABASE system effectively. To solve this problem, some systems of querying were built, such as QBIC, ADL, DBLP, Virage, Alta Vista, SIMPLYcity,...

There are two common approaches querying the images: querying the images is based on the keyword TBIR (Text-Base Image Retrieval) [2, 3] and based on the content CBIR (Content-Base Image retrieval) [2, 4, 9, 10]. The TBIR system can give the images with content that are not related to the request of querying because the nature of the keywords is independent of the content of the images. To overcome this problem, CBIR system will extract the visual attributes of the images which are needed to query, and then compare with the visual attributes of the other images which have been stored in the database. However, if the method of comparing the similarity of the content is ineffective, the results of querying will put out the images with content which are not related to the requested query. The similar assessment method of images basing on EMD distance and the method of querying images basing on the S-tree will be built in this paper.

In recent years, there has been considerable research published regarding CBIR, such as the query image system based on color histogram [2], quantization and compare similarity measure of the images based on color histogram [3, 8], the similarity of the images based on the combination of the images' colors and texture [9, 12, 13, 14, 16], the query images based on the colors [10], the query images based on the similarity of the images [4, 17], based on histogram and the texture of the images [15], using the EMD distance in image retrieval [5, 6, 18], the image indexing and retrieval technique VBA (Variable-Bin Allocation) basing on signature bitstrings and S-tree [4], color quantization in the images[11],…

In the approach of this paper will create the binary signature of an image, and also the description of the distribution of image's colors by a bitstring with given size. The content of the article will aim to query efficiently "similar images" in a large image database system. This paper will have two major parts that reducing the amount of storage space and speeding up the query image on a large database systems.

# 2 THE RELATED THEORY

## 2.1 The binary signature of the image

The binary signature of the image "the vector bit" is formed by hashing the data objects, and will have k bit 1 and (m-k) bit 0 in the bit chain [1..m], with m is the length of the binary signature. [1]

The data objects and the object of the query are encoded on the same algorithm. When the bits in the signature data object are completely covered with the bits in the query signature, then this data object is a candidate fulfilling the query. There are three cases occuring: [1] (1) the data object matches the query: every bit in the $s_q$ is covered with thebits in the signature $s_i$ of the data object (i.e., $s_q \wedge s_i = s_q$); (2) the object does not match the query (i.e., $s_q \wedge s_i \neq s_q$); (3) the signatures are compared and then give the false drop result.

The binary signature of the image is the description of the contents of the image. The binary signature method in VBA (Variable-Bin Allocation) will save over 75% and 87.5% in storage overhead when compared to GCHs (Global Color Histograms) and CCVs(Color-Coherence Vectors), respectively. [4]

Each image in the database is quantized into a fixed number of n colors: $c_1, c_2, \ldots, c_n$. Each color $c_j$ will be represented by a bitstring of length $t$, i.e., $b_1^j b_2^j \ldots b_t^j$, for $1 \leq j \leq n$, so each image will be described as a sequence of bits $S = b_1^1 b_2^1 \ldots b_t^1 \ldots b_1^n b_2^n \ldots b_t^n$, with $b_i^j = 1$ if $i = \lceil h_j \times t \rceil$, otherwise $b_i^j = 0$, where $h_j$ is the percentage of pixels of the color $c_j$ in the image. [4]

S-tree [1, 4] is a tree with many branches which are balanced, each node of an S-tree contains a number of pairs $\langle sig, next \rangle$, where *sig* is a signature and *next* is a pointer to a child node. A node root of the S-tree contains at least two pairs $\langle sig, next \rangle$ and at most M pairs $\langle sig, next \rangle$, all internal nodes in the S-tree can accommodate at least m and at most M pairs $\langle sig, next \rangle$, $1 \leq m \leq M/2$, the leaves of the S-tree will contain image signatures *sig*, along with a unique identifier *oid* for those images. The tree height for n signatures is at most: $h = \lceil \log_m n - 1 \rceil$

Each query signature will do the top-down order and can traverse many paths from root to leafs because the signature query can be match with many signatures at internal node in the S-tree.

Building the S-tree been is done by basing on inserting and splitting the node. At the beginning, the S-tree only contains a null leaf, after that each signature will be inserted into the S-tree. When the node v is full will be split into two node, at the same time the parent node $v_{parent}$ will created (if not exist) and two new signatures will be insert to node $v_{parent}$.

## 1.2 The EMD distance

The EMD distance [5, 6] is the minimum amount to be transported from one component to another. The EMD distance is based on finding the solution in the transportation problem. Setting *I* is a set of suppliers, *J* is a set of consumers, $c_{ij}$ is the transportation costs from the supplier $i \in I$ to the consumer $j \in J$, we need to find out flows $f_{ij}$ to minimize the total cost $\sum_{i \in I} \sum_{j \in J} c_{ij} f_{ij}$ [7] with the following constraints:

$$\begin{cases} f_{ij} \geq 0 & i \in I, j \in J \\ \sum_{i \in I} f_{ij} \leq y_j & j \in J \\ \sum_{j \in J} f_{ij} \leq x_i & i \in I \end{cases}$$

With $x_i$ is the provider's general ability $i \in I$, $y_j$ is the total need of the consumer $j \in J$. The feasible condition is: the total consumption total doesn't exceed the total supply : $\sum_{j \in J} y_j \leq \sum_{i \in I} x_i$. At that time, the EMD distance is defined as follow:

$$EMD(x, y) = \frac{\sum_{i \in I} \sum_{j \in J} c_{ij} f_{ij}}{\sum_{i \in I} \sum_{j \in J} f_{ij}} = \frac{\sum_{i \in I} \sum_{j \in J} c_{ij} f_{ij}}{\sum_{j \in J} y_j} \text{ s}$$

The transportation problem is used for matching the value of color histogram by defining two color histograms $H_1$ and $H_2$ are the suppliers and the consumers, respectively. The cost $c_{ij}$ is the distance between the element $i \in H_1$ and the element $j \in H_2$, $f_{ij}$ is the distribution flow from the element histogram $i \in H_1$ to the one $j \in H_2$.

# 3 Building data structures and image retrieval algorithms

## 3.1 Creating a binary signature of the image basing on the color histogram

*Step 1.* Choosing a standard color set $C = \{c_1, c_2, ..., c_n\}$ is to calculate the color histogram of the image, supposing $I$ is the image that needs to calculate color histogram. Quantification of the image $I$ in order to retain only the dominant colors $C_I = \{c_1^I, c_2^I, ..., c_{n_I}^I\}$, the color histogram vector of image $I$ is $H_I = \{h_1^I, h_2^I, ..., h_{n_I}^I\}$.

*Step 2.* The colourful histogram vector standardizes $H = \{h_1, h_2, ..., h_n\}$ the color histogram of the image $I$ in the color set $C = \{c_1, c_2, ..., c_n\}$

$$h_i = \begin{cases} 0 & c_i \notin C \cap C_I \\ \dfrac{h_j^I}{\sum_j h_J^I} & c_i \in C \cap C_I, c_i \equiv c_j^I \end{cases}$$

*Step 3.* Each color $c_j^I$ will be described into a sequence of bits with the length m: $b_1^j b_2^j, ..., b_m^j$, so the binary signature of the image $I$ will be: $b_1^1 b_2^1, ..., b_m^1 ... b_1^n b_2^n, ..., b_m^n$, in which

$$b_i^j = \begin{cases} 1 & i = \lceil h_i \times \frac{i}{m} \times 100 \rceil \\ 0 & i \neq \lceil h_i \times \frac{i}{m} \times 100 \rceil \end{cases}$$

Setting $B^j = b_1^j b_2^j ... b_m^j$, the binary signature of the image $I$ will be: $SIG = B^1 B^2 ... B^n$

## 3.2 Measurement similar image based on EMD distance

The image $I$ has the binary signature $SIG_I = B_I^1 B_I^2 ... B_I^n$, the weighted of the component $B_I^j$ will be: $w_I^j = w(B_I^j) = \sum_{i=1}^{m} (b_i^j \times \frac{i}{m} \times 100)$, with $B_I^j = b_1^j b_2^j ... b_m^j$. From those things, we have the weighted vector of the image $I$ will be: $W_I = \{w_I^1, w_I^2, ..., w_I^n\}$. $J$ is the image that we need to calculate the similarity comparing with the image $I$, so we need to minimize the colour distribution costs: $\sum_{i=1}^{n} \sum_{j=1}^{n} d_{ij} f_{ij}$, with $F = (f_{ij})$ is the matrix of color flows distribution from $c_I^i$ to $c_J^j$, and $D = (d_{ij})$ is the Euclidean distance matrix in the RGB color space from $c_I^i$ to $c_J^j$. The similarity between two images $I$ and $J$ basing on the EMD distance will minimize the value $EMD(I,J) = \min_{F=(f_{ij})} \frac{(\sum_{i=1}^{n}\sum_{j=1}^{n} d_{ij} f_{ij})}{\sum_{i=1}^{n}\sum_{j=1}^{n} f_{ij}}$ , with

$$\sum_{i=1}^{n}\sum_{j=1}^{n} f_{ij} = \min(\sum_{i=1}^{n} w_I^i, \sum_{j=1}^{n} w_J^j)$$

### 3.3 Creating similar measure image based on EMD distance

In order to reduce the storage space and increase the query speed, this report builds the signature tree S- tree to store the binary signatures of the image. Each node in the S-tree will store a set of the elements $\{\langle sig, next\rangle\}$, with $sig$ is the signature and $next$ reference pointer to the child node. The leaves will store the entries $\{\langle sig, oid\rangle\}$, with $sig$ is the signature of each image and $oid$ is the unique identification of the corresponding image. The process of creating the S-tree is based on inserting and splitting the nodes in the S-tree [1, 4]. The algorithm creating the S-tree stores the image's binary signature basing on the EMD distance as follow:

```
Input: the set of signatures S = {<sigi, oidi> | i = 1,…,n}
Output: the S-tree
Algorithm1. Gen-Stree(S, Root)
  Begin
  Step 1.
      v = Root;
      If S = ∅ then STOP;
      Else Choosing <sig, oid> ∈ S and S = S \ <sig, oid>;
      Qua bước 2;
  Step 2.
      If v is leaf then
      begin
            v = v ⊕ <sig, oid>;
            UnionSignature(v);
            If v.count > M then SplitNode(v);
            To go back Step 1;
      end
      Else
      begin
            EMD(SIG0→sig, sig) = min{EMD(SIGi→sig,sig)| SIGi ∈ v};
```

```
            v = SIG0→next;
            To go back Step 2;
      end
End.
```

The Algorithm1 will give the $sig$ signatures from the S signature file into S-tree. Each signature $sig$ will be inserted into the most appropriate leaf. If the leaf is full, the spit will be done after that the S-tree will grow in the direction of the root of the tree height. At each internal node of the S-tree, the priority will go in the similarity measure with EMD distance, this process will be approved until we can find a appropriate leaf. The traverse S-tree does not necessarily go through all paths, this will be a period of significant cost reduction in the process of finding the appropriate leaf. Therefore, each signature to insert will traverse through the height $h = \lceil \log_m n - 1 \rceil$, with $m$ is the minimum number of signatures of a node in the S-tree, $k$ is the length of each signature, every node in S-tree will have maximum $M$ signature, so the browser to find the appropriate leaf node would cost a maximum of $k \times M \times \lceil \log_m n - 1 \rceil$. However, if we find the appropriate leaf is full, we have to split node and will have the time complexity $O(M^3)$ [1]. Splitting the node basing on base of operations $\alpha - seed$, $\beta - seed$, and basing on the similarity measure EMD, is done as follows

Input: Nút cần tách v và danh sách các chữ ký cần chèn
Output: Cây S-tree sau khi thực hiện phép tách nút
Algorithm2. SplitNode(v)
Begin
  Tạo nút $v_\alpha$ và $v_\beta$ lần lượt chứa chữ ký $\alpha - seed$ và $\beta - seed$
  v = v \ { $\alpha - seed$ , $\beta - seed$ }
For ($SIG_i \in$ v)
  begin
    If (EMD(SIGi→sig, $\alpha - seed$ ) < EMD(SIGi→sig, $\beta - seed$ )) then
      $v_\alpha$ = $v_\alpha$ ⊕ SIGi;
    Else
      $v_\beta$ = $v_\beta$ ⊕ SIGi;
  end
  $s_\alpha = \bigcup SIG_i^\alpha$ , with $SIG_i^\alpha \in v_\alpha$
  $s_\beta = \bigcup SIG_i^\beta$ , with $SIG_i^\beta \in v_\beta$
  If ( $v_{parent}$ != null) then

```
        v_parent = v_parent ⊕ s_α        ;        v_parent = v_parent ⊕ s_β        ;
        UnionSignature( v_parent );
If ( v_parent.count > M ) then  SplitNode( v_parent );
    If ( v_parent = null) then
        Root = { s_α , s_β };
End.
Procedure UnionSignature( v )
Begin
    s = ⋃SIG_i , với SIG_i ∈ v ;
    If( v_parent != null) then
    begin
        SIG_v = {SIG_i | SIG_i → next = v , SIG_i ∈ v_parent} ;
        v_parent → (SIG_v → sig) = s ;
        UnionSignature( v_parent );
    end
End
```

### 3.4 The image retrieval algorithm based on S-tree and EMD distance

After storing signatures and identify of the image on the S-tree. The process of retrieval will give the signatures of the image basing on approving the S-tree and the similarity EMD. Process the same query signature image will be significantly reduced because of only browse the direction of the best measure EMD instead of having to browse in all directions in accordance with the query image signature. After finding similar image signatures, based on the identification of specific images will find images similar to the query image. Therefore, the problem is to do is to find the signature of the image and identity of the corresponding image, the query process is carried out as follows

```
Input: the query signature sig and the S-tree
Output: Set of image's binary signatures and Set of unique identify references to image.
Algorithm3. Search-Image-Sig(sig, S-tree)
Begin
    v = root;
SIGOUT = ∅;
    Stack = ∅;
    Push(Stack, v);
    while(not Empty(Stack)) do
    begin
```

```
            v = Pop(Stack);
            If(v is not Leaf) then
  begin
                  For(SIGi ∈ v and SIGi→sig ∧ sig = sig)
do
                        EMD(SIG0→sig,        sig)        =
min{EMD(SIGi→sig, sig)| SIGi ∈ v};
                  Push(Stack, SIG0 → next);
            end
            Else
                  SIGOUT = SIGOUT ∪ {<SIGi → sig, oidi>
| SIGi ∈ v};
  end
  return SIGOUT;
  End.
```

Since the S-tree has many tree multiple balanced branches. In addition to, at each node of the tree will be travesed in the best similirity measure direction next with EMD, which will cost up to browse the S-tree is $h = \lceil \log_m n - 1 \rceil$. Search process on is done similar to traverse the S-tree, so the cost of the query process on S-tree is $k \times M \times \lceil \log_m n - 1 \rceil$, with k is the length of each signature, m is the minimum number of signatures, M is the maximum number of signatures of a node in the S-tree.

# 4 Experimental

## 4.1 Model application

Based on the theory and algorithms proposed above, this paper carried out empirical application built to evaluate and verify the theoretical basis proposed. The empirical application build process consists of two phases, the first phase will perform preprocessing to convert image database becomes binary signature and put into signature S-tree based on the relative measureself EMD. The second phase will execute the query process, corresponding to a query image needs to be converted to the binary signature and the query will be performed on the S-tree based on the similarity measure EMD. After the signature of the similar image, retrieves specific images and sorted by priority order of the similarity measure EMD.

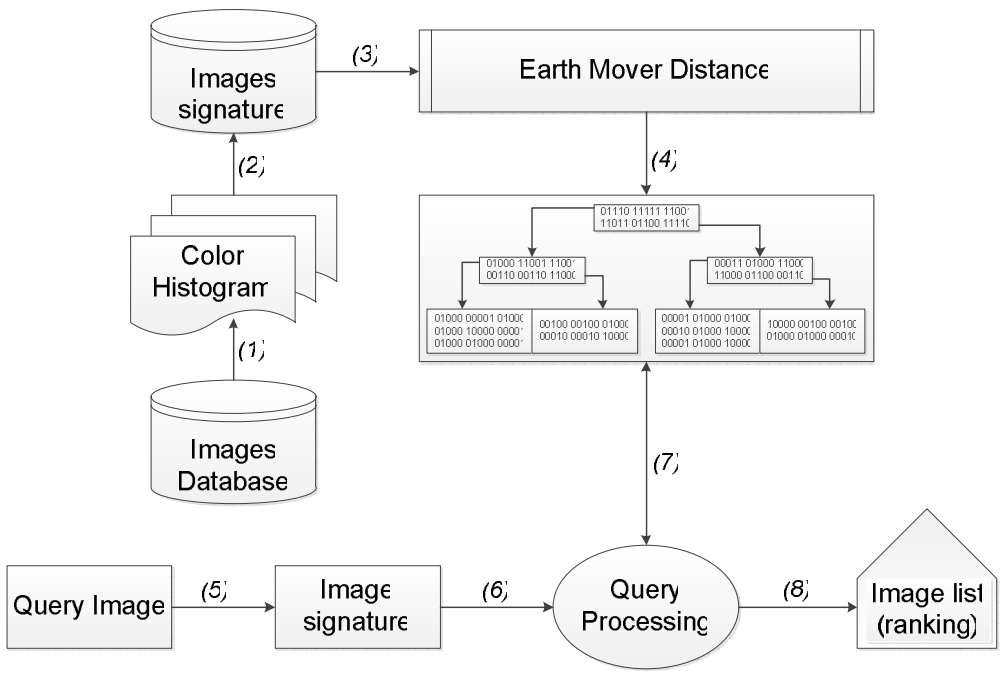

**Fig. 1.** Model image retrieval system

Phase 1: Perform preprocessing

Step 1. Quantized images in the database and converted to a color histogram.

Step 2. Convert the color histogram of the image in the form of binary signatures.

Step 3. Respectively calculate the EMD distance of the signature images and insert the signature image into the S-tree

Phase 2: Implementation query

Step 1. For each query image, will calculate the color histogram and converted into binary signatures.

Step 2. Perform binary signature query on S-tree consists of the signature images, it is possible to find similar images at the leafs of the S-tree through the EMD measure.

Step 3. After images similar conduct arrangement similar level from high to low and make title match after images arranged on the basis of similarity EMD distance.

### 4.2 The experimental results

Each image will calculate the color histogram based on 16 color range is divided on the HSV color space include: BLACK, SILVER, WHITE, GRAY, RED, ORANGE, YELLOW, LIME GREEN, TURQUOISE, CYAN, OCEAN, BLUE, VIOLET, MAGENTA, RASPBERRY

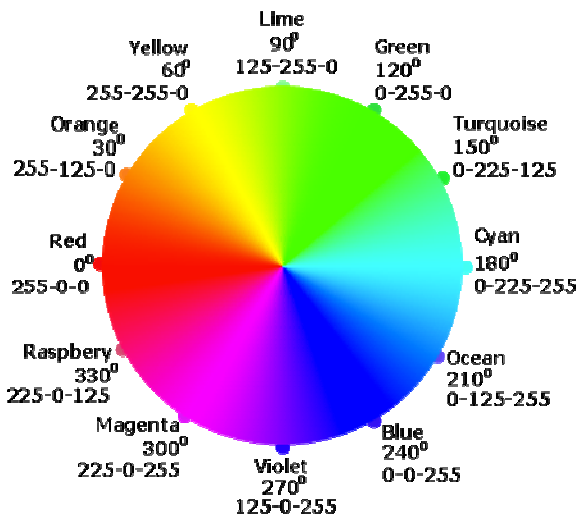

**Fig. 2.** Standard color range to calculate the color histogram

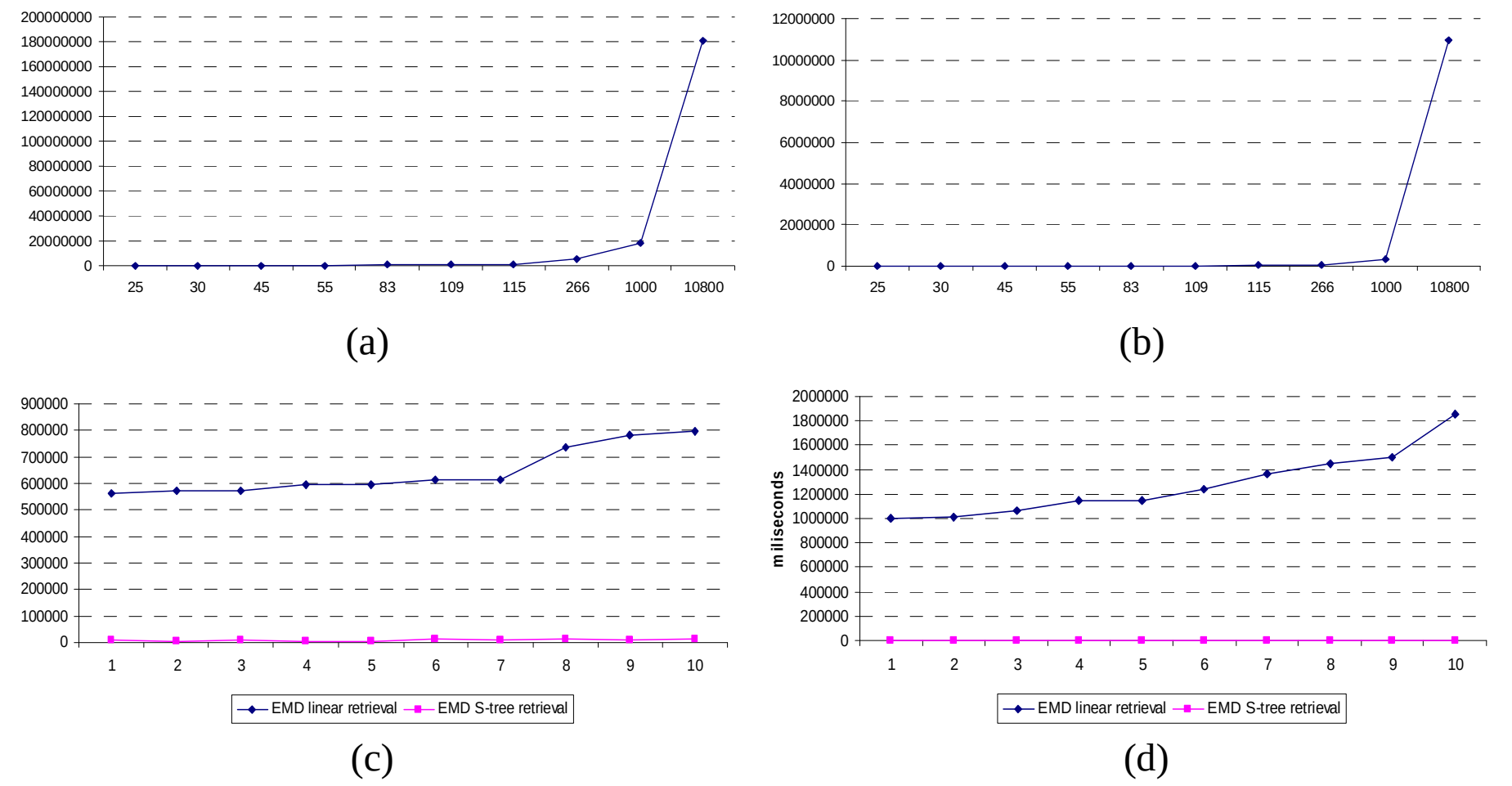

**Fig. 3.** (a) Number of comparisons to create S-tree. (b) The time (miliseconds) to create S-tree. (c) Number of comparisons to query image in database over 10,000 images. (d) The time (miliseconds) to query image in database over 10,000 images.

# 5 Conclusion

The paper created the algorithms in order to speed up the retrieval similar image based on the image's binary signatures, thence the paper has designed and implemented the image retrieval model on CBIR’s content. As the experimental showed that the it is takes a long time to create S-tree from the image’s binary signature, but the retrieval image relied on S-tree will be a lot faster than a linear search method based on EMD. However, when using EMD to calculate the distribution of the image’s colors will result in inaccurate in the case of images with the same percentage of color pixels but the color distribution location does not correspond to each other. The next development will assess the similarity of the image through EMD distance with location distribution of the percentage of color pixels and compare objects in the contents of image to increase accuracy when querying the similar images.

# References

1. Yangjun Chen, Yibin Chen: On the Signature Tree Construction and Analysis, IEEE Trans. Knowl. Data Eng., vol. 18, no. 9, pp.1207--1224 (2006)
2. Neetu Sharma. S, Paresh Rawat. S, Jaikaran Singh. S,: Efficient CBIR Using Color Histogram Processing, Signal & Image Processing: An International Journal, vol. 2, no. 1, pp. 94--112 (2011)
3. Fazal Malik, Baharum Bin Baharudin: Feature Analysis of Quantized Histogram Color Features for Content-Based Image Retrieval Based on Laplacian Filter, International Conference on System Engineering and Modeling, vol. 34, pp. 44--49 (2012)

4. Mario A. Nascimento, Eleni Tousidou, Vishal Chitkara, Yannis Manolopoulos: Image indexing and retrieval using signature trees, Data & Knowledge Engineering, vol. 43, no. 1, pp. 57--77 (2002)
5. Yossi Rubner, Carlo Tomasi, and Leonidas J. Guibas: A Metric for Distributions with Applications to Image Databases, Proceedings of the IEEE International Conference on Computer Vision, Bombay, India, pp. 59--66, 4-7 Jan 1998
6. Bahri abdelkhalak, Hamid zouaki: EMD Similarity Measure and Metric Access Method using EMD Lower Bound, International Journal of Computer Science & Emerging Technology, vol. 2, no. 6, pp. 323--332 (2011)
7. K. Konstantinidis, A. Gasteratos, I. Andreadis: Image retrieval based on fuzzy color histogram processing, Science Direct, Optics Communications, vol. 248, issue 4-6, pp. 375--386 (2005).
8. Ibrahim S. I. Abuhaiba, Ruba A. A. Salamah: Efficient Global and Region Content Based Image Retrieval, International Journal of Image, Graphics and Signal Processing, vol. 4, no. 5, pp. 38--46 (2012)
9. Rahul Mehta, Nishchol Mishra, Sanjeev Sharma: Color - Texture based Image Retrieval System, International Journal of Computer Applications, vol. 24, no. 5, pp. 24--29 (2011)
10. Gunja Varshney, Uma Soni: Color-Based Image Retrieval in Image Database System, International Journal of Soft Computing and Engineering, vol. 1, no. 5, pp.31--35 (2011)
11. Shyi-Chyi Cheng, Chen-Kuei Yang: A fast and novel technique for color quantization using reduction of color space dimensionality, Pattern Recognizer Letter, vol. 22, issue 8, pp. 845--956 (2001)
12. Manimala Singha, K.Hemachandran: Content Based Image Retrieval using Color and Textual, Signal & Image Processing: An International Journal, vol. 3, no. 1, pp. 39--57 (2012)
13. Sridhar, Gowri: Color and Texture Based Image Retrieval, ARPN Journal of Systems and Software, vol. 2, no.1, pp. 1--6 (2012)
14. T. Mehyar, J. O. Atoum: An Enhancement on Content-Based Image Retrieval using Color and Texture Features, Jour. of Emerging Trends in Computer and Information Science, vol. 3, no. 4, pp.488--496 (2012)
15. Ch. Kavitha, M. Babu Rao, B.Prabhakara Rao, A.Govardhan: Image Retrieval based on Local Histogram and Texture Features, International Journal of Computer Science and Information Technology, vol. 2, no. 2, pp.741--746 (2011)
16. Chuen-Horng Lin, Rong-Tai Chen, Yung-Kuan Chan: A smart content-based image retrieval system based on color and texture feature, Image and Vision Computing, vol. 27, no, 6, pp. 658--665 (2009)
17. J. Yu, J. Amores, N. Sebe, P. Radeva, Q. Tian: Distance Learning for Similarity Estimation, IEEE Transactions On Pattern Analysis and Machine Intelligence, vol. 30, no. 3, pp. 451--462 (2008)
18. Thomas Hurtut, Yann Gousseau, Francis Schmitt:Adaptive Image retrieval based on the spatial organization of colors, Computer Vision and Image Understanding, vol. 112, no. 2, pp. 101--113 (2008)

## Appendix:

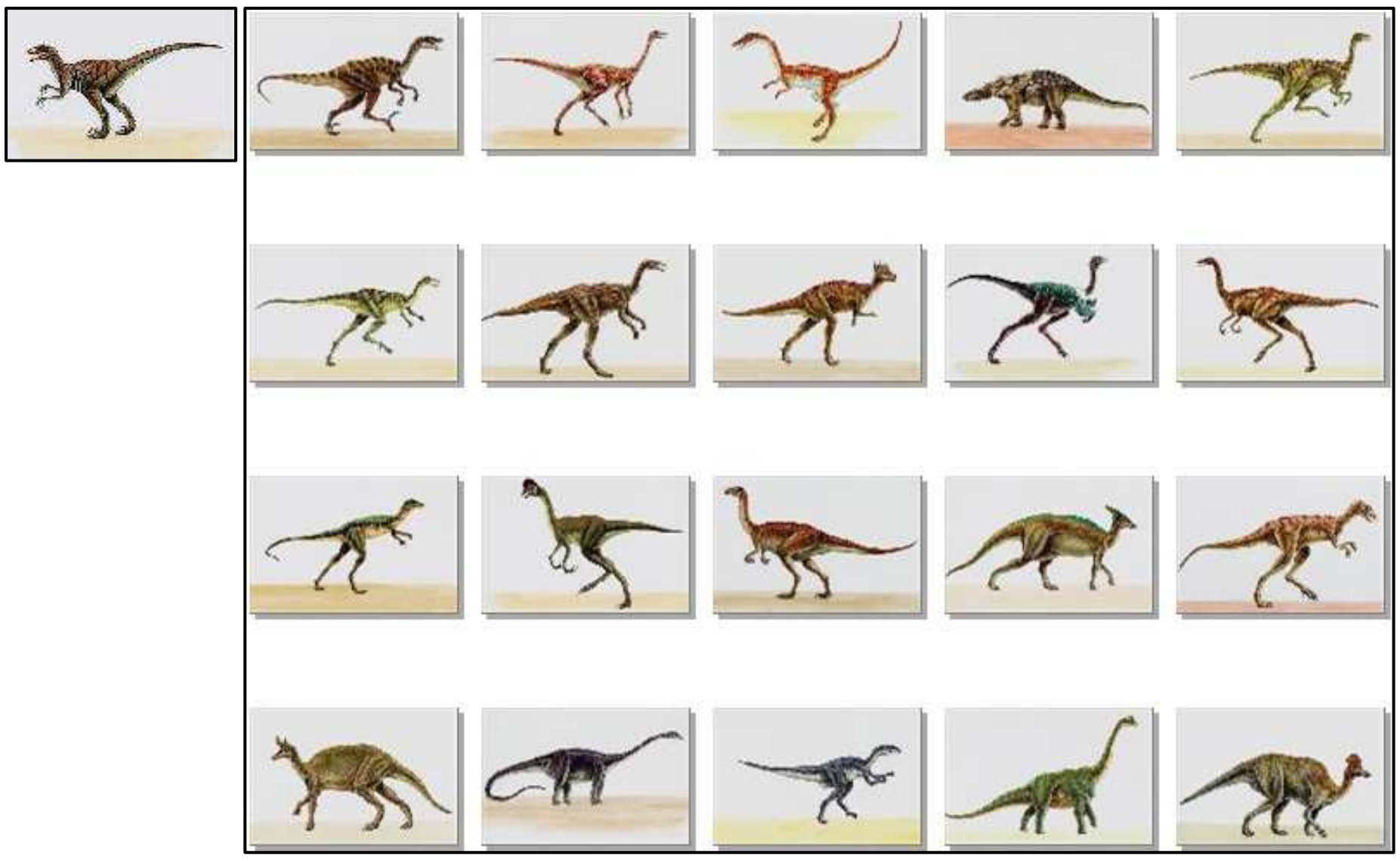

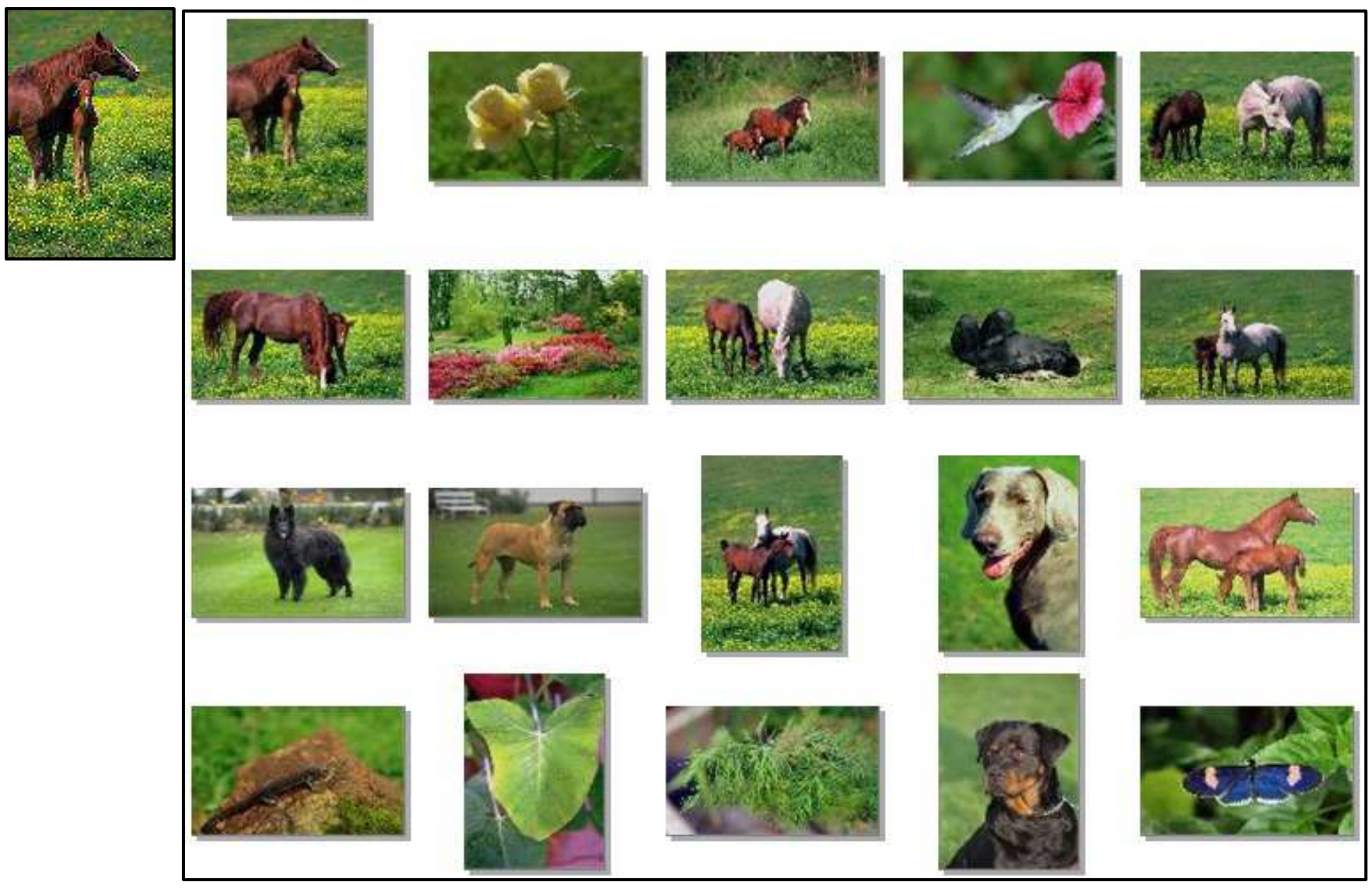

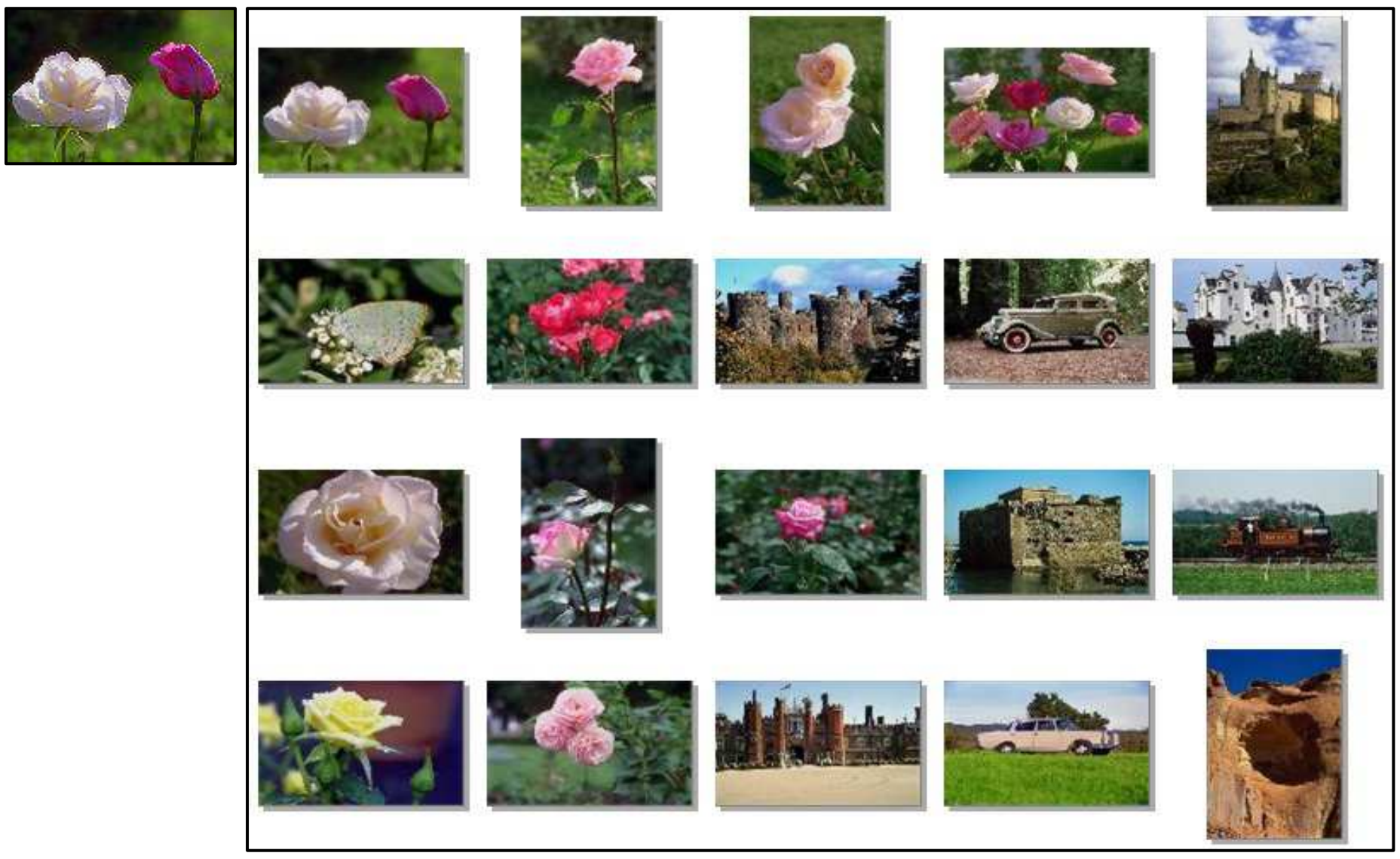

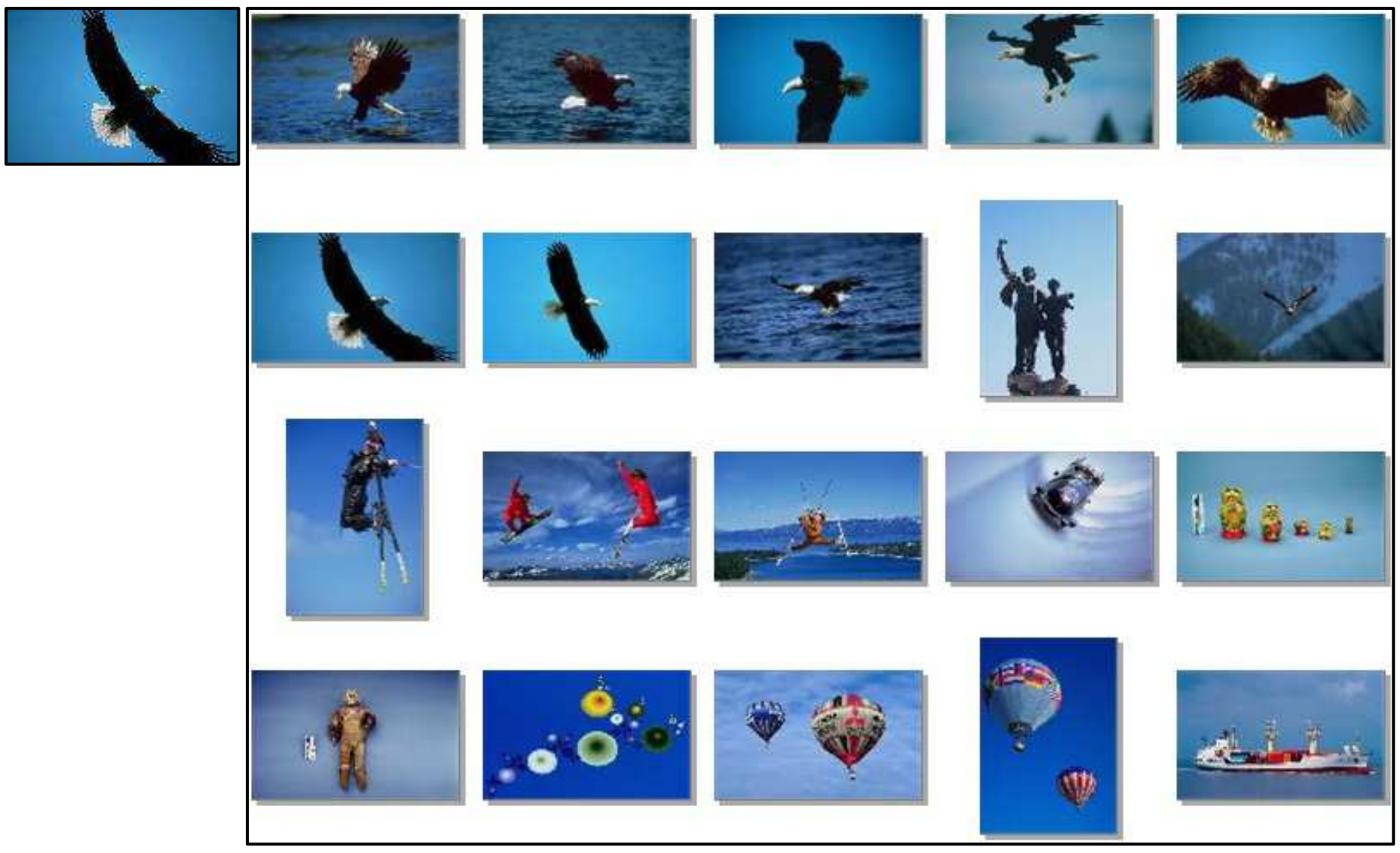

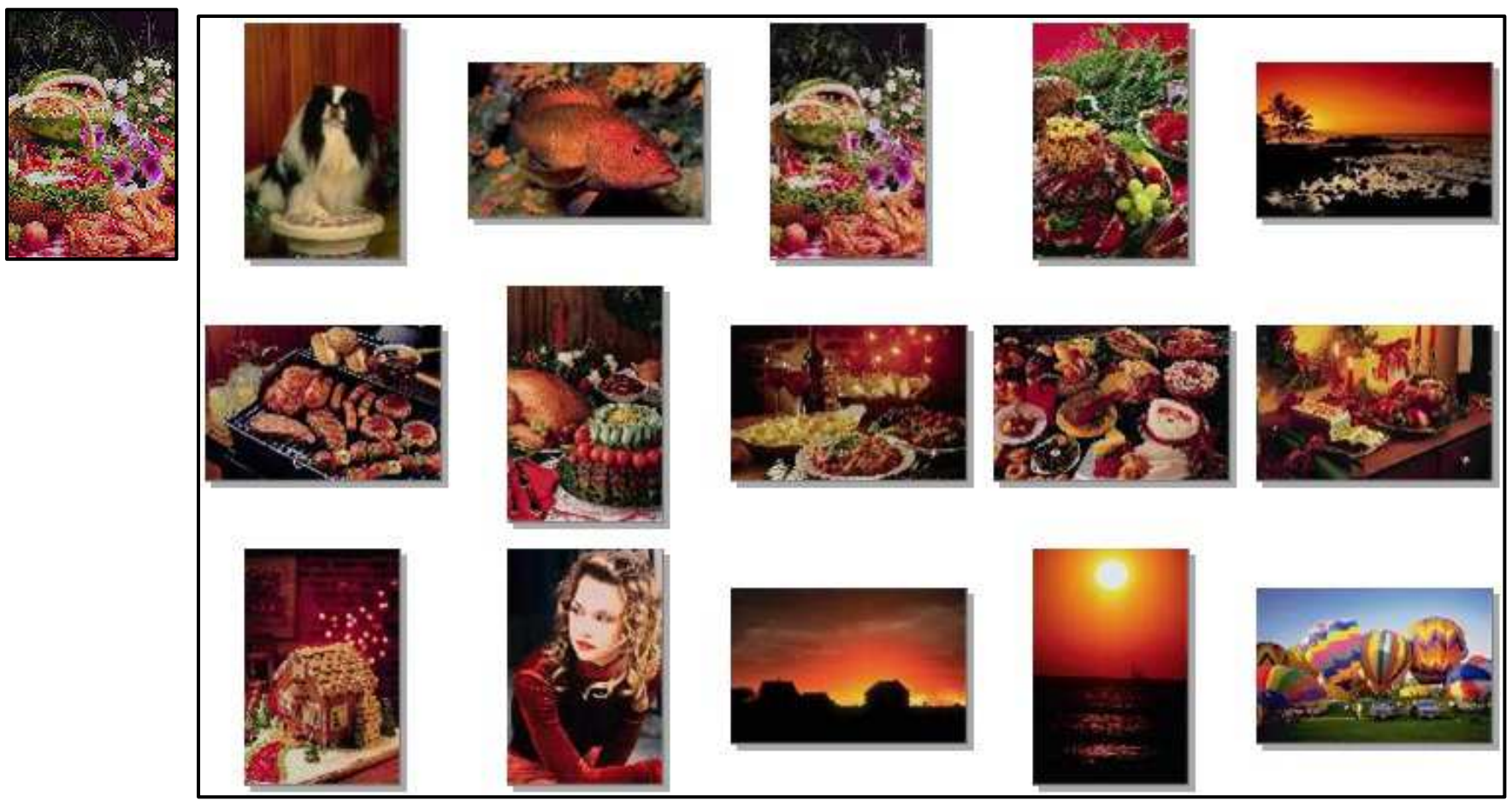

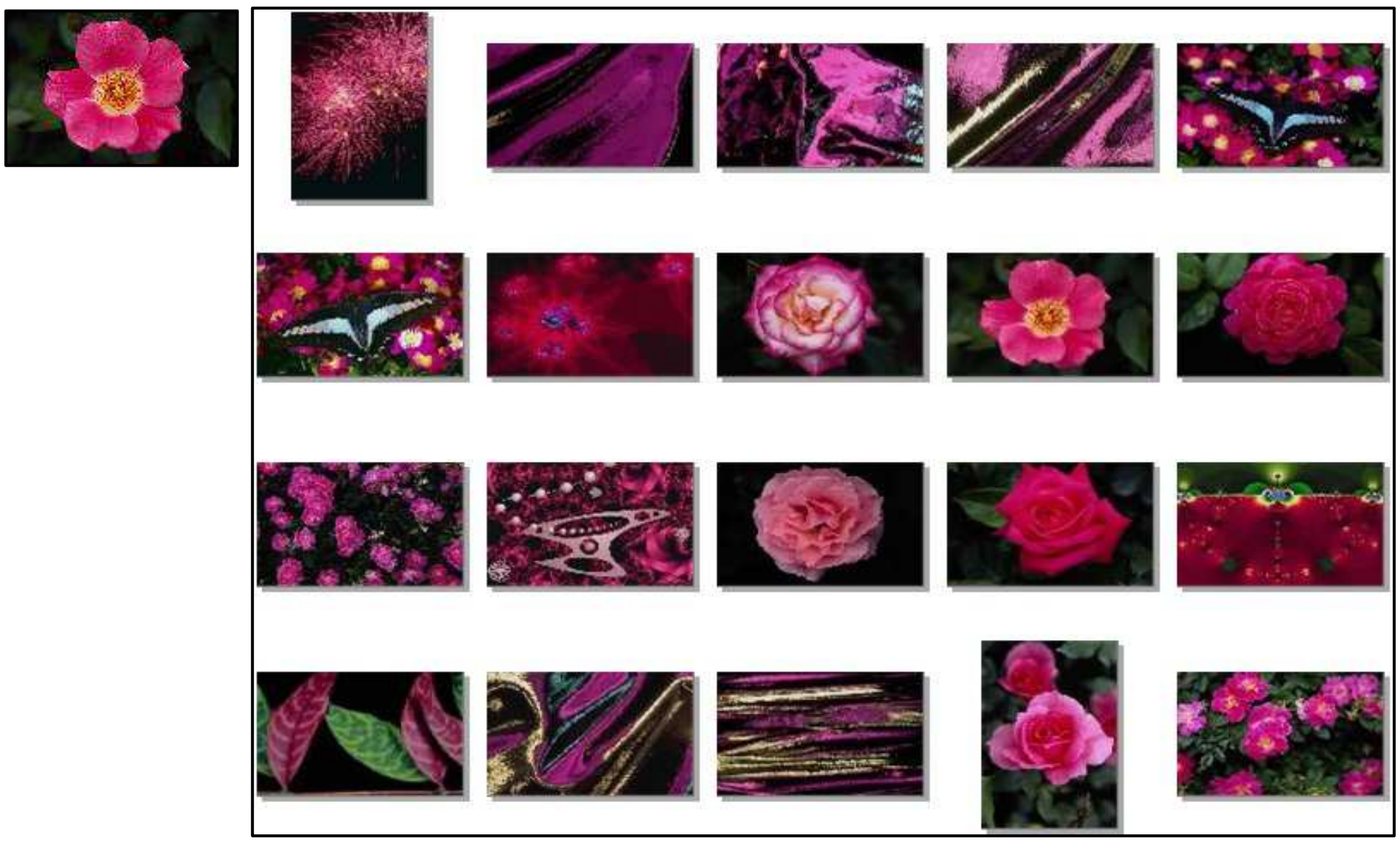